\documentclass[conference]{IEEEtran}
\usepackage[utf8]{inputenc}
\usepackage[T1]{fontenc}
\usepackage[english]{babel}

\usepackage{cite}
\usepackage{amsmath,amssymb,amsfonts}
\usepackage{algorithmic}
\usepackage{graphicx}
\usepackage{textcomp}
\usepackage{xcolor}
\usepackage{tikz}
\usepackage[hidelinks]{hyperref}

\newif\ifanonymous
\anonymousfalse

\usetikzlibrary{arrows.meta, positioning, fit}

\def\BibTeX{{\rm B\kern-.05em{\sc i\kern-.025em b}\kern-.08em
    T\kern-.1667em\lower.7ex\hbox{E}\kern-.125emX}}
\begin{document}

\title{Toward Modeling Player-Specific Chess Behaviors}

\ifanonymous
\else
	\author{\IEEEauthorblockN{Loris Sogliuzzo}
		\IEEEauthorblockA{\textit{ICTEAM, UCLouvain} \\
			Louvain-la-Neuve, Belgium \\
		}
		\and
		\IEEEauthorblockN{Aloïs Rautureau}
		\IEEEauthorblockA{\textit{ICTEAM, UCLouvain} \\
			Louvain-la-Neuve, Belgium \\
		}
		\and
		\IEEEauthorblockN{Eric Piette}
		\IEEEauthorblockA{\textit{ICTEAM, UCLouvain} \\
			Louvain-la-Neuve, Belgium \\
		}
	}
\fi

\maketitle

\begin{abstract}
	While artificial intelligence has achieved superhuman performance in chess, developing models that accurately emulate the individualized decision-making styles of human players remains a significant challenge. Existing human-like chess models capture general population behaviors based on skill levels but fail to reproduce the behavioral characteristics of specific historical champions. Furthermore, the standard evaluation metric, move accuracy, inherently penalizes natural human variance and ignores long-term behavioral consistency, leading to an incomplete assessment of stylistic fidelity. To address these limitations, an architecture is proposed that adapts the unified Maia-2 model to champion-specific embeddings, further enhanced by the integration of a limited Monte Carlo Tree Search (MCTS) process to enrich tactical exploration during move selection. To robustly evaluate this approach, a novel behavioral metric based on the Jensen-Shannon divergence is introduced. By compressing high-dimensional board representations into a latent space using an AutoEncoder and Uniform Manifold Approximation and Projection (UMAP), move distributions are discretized on a common grid to compare behavioral similarities. Results across 16 historical world champions indicate that while integrating MCTS decreases standard move accuracy, it improves stylistic alignment according to the proposed metric, substantially reducing the average Jensen-Shannon divergence. Ultimately, the proposed metric successfully discriminates between individual players and provides promising evidence toward more comprehensive evaluations of behavioral alignment between players and AI models.
\end{abstract}


\section{Introduction}

Historically, chess has been used as a sandbox for developing powerful AI systems. More recently, research has shifted toward modeling human-like play by reproducing characteristic behaviors, mistakes, and decision-making patterns. Maia \cite{McIlroy_Young_2020} and Maia-2 \cite{tang2024maia2unifiedmodelhumanai} successfully capture population-level behaviors associated with different Elo ranges, but modeling the individualized behaviors of specific historical champions remains an open challenge. More broadly, this work aligns with one of the central research directions of the COST Action GameTable \cite{gametable}, namely the modeling of human behavioral patterns and player-specific decision-making in complex game environments. Related efforts in this direction have also recently explored human-like modeling in broader general game-playing settings \cite{cogniplay}.

The objective of this ongoing research is to develop agents capable of reproducing player-specific chess behaviors. To this end, Maia-2 is adapted through champion-specific embeddings, and a behavioral evaluation framework based on the Jensen-Shannon divergence is introduced to compare player and model board-transition distributions.

\section{Related Work}
The attempt to model human chess play was significantly advanced by Maia, an agent based on the AlphaZero \cite{silver2017masteringchessshogiselfplay} architecture but trained strictly on human games without the use of Monte Carlo Tree Search (MCTS) \cite{10.1007/978-3-540-75538-8_7}. This work was built upon by Maia-2, which introduced a skill-aware attention mechanism to capture the skill level of players based on their Elo ratings. Crucially, rather than training separate models, Maia-2 dynamically adapts to different skill levels within a single unified model by using categorical population embeddings. However, Maia and Maia-2 were not designed to capture the playstyle of specific players, but rather to capture the general playstyle of populations partitioned based on skill level.

To address this limitation and move toward true individualization, Maia4All \cite{tang2025learningimitatelessefficient} extends this architecture by generalizing population embeddings into individual embeddings. It attempts to model specific individuals through a two-stage framework. It first enriches the base model using a pool of prototype players (requiring around 2,500 games per prototype). Subsequently, it initializes a new embedding for a target player using a Prototype Matching Network (PMN) that learns to match the unseen player to the prototypes. Finally, this initialized embedding is fine-tuned to capture the target player's specific playstyle using as few as 20 games.

Current evaluations of human-like chess agents primarily rely on move accuracy, a deterministic metric that fails to capture the natural variability of human decision-making. Alternative behavioral metrics are thus needed to evaluate stylistic alignment more robustly.

\section{Methods}

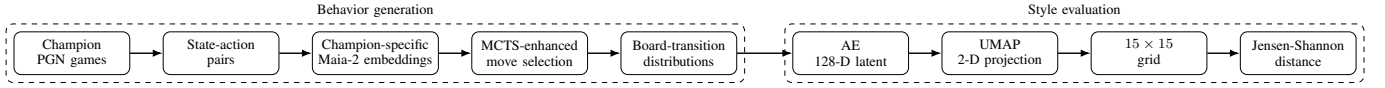
\begin{figure*}[!t]
	\centering
	\resizebox{\textwidth}{!}{
		\begin{tikzpicture}[
			node distance=0.65cm,
			box/.style={
					rectangle,
					rounded corners,
					draw,
					align=center,
					minimum width=2.3cm,
					minimum height=0.85cm,
					font=\footnotesize
				},
			arrow/.style={-{Latex[length=2mm]}, thick}
			]

			\node[box] (pgn) {Champion\\PGN games};
			\node[box, right=of pgn] (pairs) {State-action\\pairs};
			\node[box, right=of pairs] (ft) {Champion-specific\\Maia-2 embeddings};
			\node[box, right=of ft] (mcts) {MCTS-enhanced\\move selection};
			\node[box, right=of mcts] (transitions) {Board-transition\\distributions};

			\node[box, right=1.1cm of transitions] (ae) {AE\\128-D latent};
			\node[box, right=of ae] (umap) {UMAP\\2-D projection};
			\node[box, right=of umap] (bins) {$15\times 15$\\grid};
			\node[box, right=of bins] (jsd) {Jensen-Shannon\\distance};

			\draw[arrow] (pgn) -- (pairs);
			\draw[arrow] (pairs) -- (ft);
			\draw[arrow] (ft) -- (mcts);
			\draw[arrow] (mcts) -- (transitions);
			\draw[arrow] (transitions) -- (ae);
			\draw[arrow] (ae) -- (umap);
			\draw[arrow] (umap) -- (bins);
			\draw[arrow] (bins) -- (jsd);

			\node[draw, dashed, rounded corners,
				fit=(pgn)(pairs)(ft)(mcts)(transitions),
				inner sep=0.15cm,
				label={[font=\footnotesize]above:Behavior generation}] {};

			\node[draw, dashed, rounded corners,
				fit=(ae)(umap)(bins)(jsd),
				inner sep=0.15cm,
				label={[font=\footnotesize]above:Style evaluation}] {};

		\end{tikzpicture}
	}
	\caption{Overview of the proposed pipeline. Champion games are used to fine-tune player-specific Maia-2 embeddings and optionally guide MCTS move selection. Player and model board transitions are then projected, discretized, and compared using the Jensen-Shannon distance.}
	\label{fig:pipeline}
\end{figure*}

The proposed approach consists of two complementary components: a behavior generation pipeline, which adapts Maia-2 to individual champions and enhances move selection with Monte Carlo Tree Search (MCTS), and a style evaluation pipeline, which compares player and model behaviors through projected board-transition distributions. An overview of the complete pipeline is shown in Fig.~\ref{fig:pipeline}.

\subsection{Data Preparation}
This work focuses on 16 chess champions from the 20th century to ensure sufficient game availability for training and evaluation according to the data requirements identified by Maia4All \cite{tang2025learningimitatelessefficient}. Games were collected from Chessgames.com and converted into state-action pairs. Because the selected players span different historical eras, part of the measured stylistic differences may also reflect broader evolutions in chess theory and opening preparation.

\subsection{Model Architecture}
To capture the specific playstyle of each champion, the embedding matrix of Maia-2 is extended by creating one new embedding per player in our set based on the `$>2000$' Elo embedding of Maia-2, the highest Elo embedding that it possesses. During training, the entire Maia-2 model is frozen to preserve the general chess knowledge it has acquired, and only these new embeddings are fine-tuned to fit the specific playstyle of each champion. This approach allows us to leverage the general chess knowledge of Maia-2 while adapting it to capture the unique style of each champion. The new embeddings were fine-tuned for 100 epochs using the Adam optimizer with a learning rate of $5 \times 10^{-4}$, a batch size of 2048, and a Cross-Entropy loss function. These parameters were determined empirically by monitoring convergence to ensure stable learning.

To enrich tactical exploration, a Monte Carlo Tree Search (MCTS) is implemented at the output of the fine-tuned Maia-2 model. The MCTS is guided by a Predictor Upper Confidence bound applied to Trees (PUCT), which explores the action $a$ in state $s$ that maximizes (\ref{eq:puct}):
\begin{equation}
	PUCT(s, a) = Q(s, a) + c \cdot P(s, a) \frac{\sqrt{N(s)}}{1 + N(s, a)}
	\label{eq:puct}
\end{equation}
where $Q(s, a)$ is the expected reward for action $a$ from state $s$, $P(s, a)$ is the prior probability of selecting action $a$ as predicted by the policy head of the Maia-2 model, $N(s, a)$ is the visit count of the state-action pair, $N(s)$ is the total visit count of the parent node, and $c$ is a hyperparameter controlling the degree of exploration. The expected reward $Q(s, a)$ is calculated using the value of each expanded state, which is given by the value head of Maia-2 if it is not a terminal state, and is set to 1, -1, or 0 depending on a win, loss, or draw.

To prevent excessive exploration from distorting player-specific behaviors, a pruning mechanism removes nodes whose prior probability falls below a fixed threshold $\eta$. Additionally, the number of simulations is limited to reduce the influence of search on the final decision process.

\subsection{Evaluation Metrics}

The main objective of this research is to develop an agent capable of reproducing the behavioral characteristics of a specific chess champion. In this work, style is operationalized as the similarity between the state-transition distributions induced by a player and those generated by the model. Unlike standard move accuracy, which evaluates discrete decisions, the proposed approach compares behavioral distributions in order to better capture the variability of human decision-making.

To compare these distributions, the Jensen-Shannon divergence (JSD) \cite{61115} is utilized. Unlike the Kullback-Leibler divergence (KL) \cite{1320776d-9e76-337e-a755-73010b6e4b64}, the JSD is symmetric and bounded between 0 and 1, making it easier to interpret. A value of 0 indicates identical distributions and therefore perfect stylistic  alignment. Additionally, the square root of the JSD defines a metric space, enabling meaningful distance comparisons between behavioral distributions.

Formally, let $P$ and $Q$ denote the empirical distributions of board transitions induced by the player and the model over a set of sampled positions, respectively. The JSD is defined as:
\begin{equation}
	JSD(P \parallel Q) = \frac{1}{2} D_{KL}(P \parallel M) + \frac{1}{2} D_{KL}(Q \parallel M)
	\label{eq:jsd}
\end{equation}
where $M = \frac{1}{2}(P + Q)$ is the average distribution, and $D_{KL}$ is the KL divergence. The resulting distance metric is given by:
\begin{equation}
	d_{JS}(P, Q) = \sqrt{JSD(P \parallel Q)}
	\label{eq:js_distance}
\end{equation}

Implementing this metric directly in the raw input space of the model is not feasible due to the high dimensionality and sparsity of the data. The input representation is an $8 \times 8 \times 18$ tensor encoding board occupancy, side-to-move, castling rights, and en passant status. By concatenating the flattened board states before and after each move, sparse transition vectors of 2304 dimensions are obtained. Direct divergence estimation in this high-dimensional sparse space is statistically unreliable. Moreover, the evaluated models do not necessarily expose directly comparable calibrated probability distributions across configurations, particularly after integrating search.

To address this issue, a dimensionality reduction pipeline is employed. A fully connected AutoEncoder (AE) \cite{doi:10.1126/science.1127647} with symmetric architecture ($2304 \rightarrow 1024 \rightarrow 512 \rightarrow 256 \rightarrow 128$) is trained to project board transitions into a compact 128-dimensional latent space. The AE is optimized using Adam with a learning rate of $10^{-3}$ for 10 epochs and a batch size of 1024. Since this representation remains too high-dimensional for efficient distribution estimation using the JSD, Uniform Manifold Approximation and Projection (UMAP) \cite{mcinnes2020umapuniformmanifoldapproximation} is subsequently applied to obtain a 2D manifold preserving both local and global structure.

Finally, calculating the JSD over continuous 2D coordinates requires computationally expensive density estimation procedures. To simplify this process, the projected manifold is discretized into a regular $15 \times 15$ grid bounded by the minimum and maximum projected coordinates. This resolution was empirically selected to balance descriptive power and statistical density while avoiding excessive sparsity in the bins. The JSD is then computed over these discrete bin distributions.

Although this metric provides promising results for distinguishing player-specific behaviors, it should be noted that part of the measured similarity may also reflect distributional smoothing effects introduced by the search procedure and dimensionality reduction pipeline, rather than stylistic fidelity alone.


\section{Results and Discussion}
\ifanonymous
\else
	The experimental results discussed in this section were obtained using a dedicated codebase, which is freely available at \url{https://github.com/SogliuzzoL/Toward-Modeling-Player-Specific-Chess-Behaviors.git}. All corresponding tests and model evaluations were performed on a cluster node featuring an Nvidia RTX 6000 ADA GPU with 48 GB of GDDR6 memory, a 32 core CPU based on the AMD EPYC Genoa architecture, and 128 GB of system RAM.
\fi

\subsection{Move Accuracy Evaluation}
To establish a baseline, the performance of the original Maia-2 model with the `$>2000$' Elo rating, the fine-tuned Maia-2 model (Maia-2 FT), and the fine-tuned Maia-2 model with MCTS (Maia-2 FT + MCTS) is evaluated in terms of move accuracy. For the MCTS configuration, $100$ simulations are utilized, and parameters are set to $c = 1.5$ and $\eta = 0.01$. Evaluation is performed independently for each player on a held-out test split representing $20\%$ of the games. For move-accuracy evaluation, Maia-2 and Maia-2 FT select the highest-probability policy action, whereas Maia-2 FT + MCTS selects the most visited action.

The results, presented in Table \ref{tab:move_accuracy}, indicate that the fine-tuning process leads to an improvement of $0.5\%$ in mean accuracy compared to the original Maia-2 model. The introduction of MCTS significantly reduces this measure by $18.7\%$ on average, likely due to the broader exploration induced by MCTS which increases behavioral diversity. This further highlights the limitation of move accuracy for evaluating style-oriented agents.

\begin{table}[!t]
	\renewcommand{\arraystretch}{1.3}
	\caption{Move-accuracy of the different models on the test set. Values following the $\pm$ indicate the standard deviation.}
	\label{tab:move_accuracy}
	\centering
	\scriptsize
	\setlength{\tabcolsep}{3pt}
	\begin{tabular}{l c c c}
		\hline
		\bfseries Player  & \bfseries Maia-2           & \bfseries Maia-2 FT        & \bfseries Maia-2 FT + MCTS \\
		\hline\hline
		Alekhine          & $45.8\%\pm0.2\%$           & 47.2\% $\pm0.2\%$          & 27.0\% $\pm0.2\%$          \\
		Anand             & $45.3\%\pm0.1\%$           & 46.4\% $\pm0.1\%$          & 26.7\% $\pm0.1\%$          \\
		Andersson         & $44.4\%\pm0.2\%$           & 44.6\% $\pm0.2\%$          & 26.8\% $\pm0.2\%$          \\
		Beliavsky         & $44.0\%\pm0.2\%$           & 44.6\% $\pm0.2\%$          & 25.3\% $\pm0.1\%$          \\
		Capablanca        & $48.6\%\pm0.3\%$           & 48.5\% $\pm0.3\%$          & 28.7\% $\pm0.2\%$          \\
		Fischer           & $48.5\%\pm0.3\%$           & 49.3\% $\pm0.3\%$          & 28.2\% $\pm0.2\%$          \\
		Ivanchuk          & $44.3\%\pm0.1\%$           & 44.5\% $\pm0.1\%$          & 25.3\% $\pm0.1\%$          \\
		Karpov            & $44.2\%\pm0.1\%$           & 45.0\% $\pm0.1\%$          & 25.7\% $\pm0.1\%$          \\
		Kasparov          & $45.2\%\pm0.2\%$           & 45.3\% $\pm0.2\%$          & 26.3\% $\pm0.2\%$          \\
		Korchnoi          & $44.3\%\pm0.1\%$           & 44.2\% $\pm0.1\%$          & 25.3\% $\pm0.1\%$          \\
		Larsen            & $43.8\%\pm0.2\%$           & 43.5\% $\pm0.2\%$          & 25.0\% $\pm0.1\%$          \\
		Petrosian         & $43.3\%\pm0.2\%$           & 43.4\% $\pm0.2\%$          & 25.9\% $\pm0.2\%$          \\
		Portisch          & $43.6\%\pm0.2\%$           & 44.3\% $\pm0.2\%$          & 25.6\% $\pm0.1\%$          \\
		Short             & $44.2\%\pm0.2\%$           & 44.7\% $\pm0.2\%$          & 25.6\% $\pm0.1\%$          \\
		Tal               & $46.3\%\pm0.2\%$           & 46.6\% $\pm0.2\%$          & 27.1\% $\pm0.2\%$          \\
		Timman            & $44.6\%\pm0.1\%$           & 45.0\% $\pm0.1\%$          & 25.7\% $\pm0.1\%$          \\
		\hline
		\bfseries Average & \bfseries $45.0\%\pm0.2\%$ & \bfseries $45.5\%\pm0.2\%$ & \bfseries $26.3\%\pm0.2\%$ \\
		\hline
	\end{tabular}
\end{table}

\subsection{Jensen-Shannon Divergence Evaluation}
\begin{figure}[!ht]
	\centering
	\includegraphics[width=\linewidth]{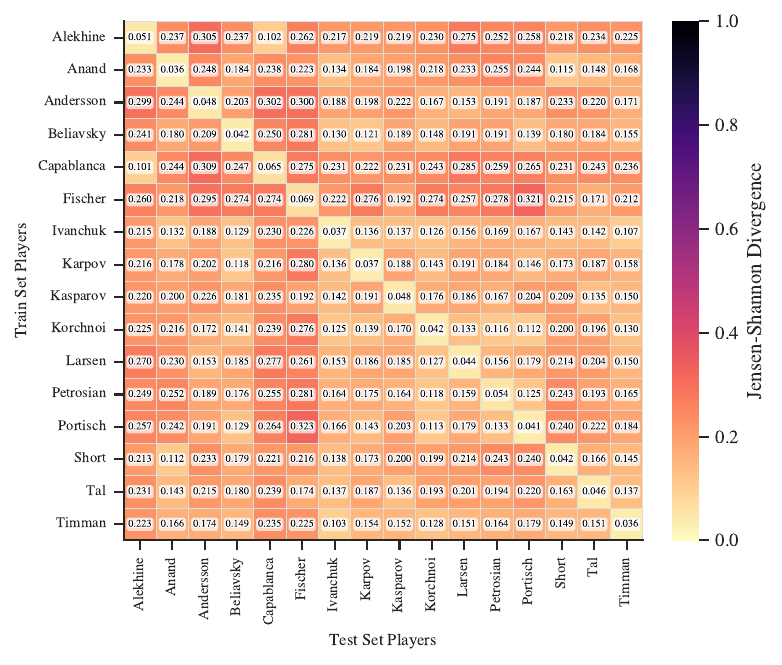}
	\caption{Heatmap representing the Jensen-Shannon divergence between the train and test sets for each player.}
	\label{fig:heatmap_train_test}
\end{figure}

To validate the relevance of the new metric based on the Jensen-Shannon divergence, the divergence between the training set and the test set of each player was first computed. As shown in Fig. \ref{fig:heatmap_train_test}, the divergence between a player's training set and test set is very low for each player compared to the divergence between the training set of one player and the test set of another, with a maximum divergence of $0.069$ for the same player and a minimum of $0.101$ for different players. These results indicate that the proposed metric consistently discriminates between player-specific behavioral distributions.

Some player pairs exhibit lower divergences than others, suggesting partial similarities in their induced behavioral distributions. For instance, Timman and Ivanchuk appear relatively close, whereas Portisch and Fischer remain substantially separated. Clusters of relatively low divergence appear between several players from similar competitive eras, suggesting that the proposed metric may capture broader behavioral regularities beyond individual identity.

\subsection{Stylistic Alignment Results}

Finally, the Jensen-Shannon divergence is applied to evaluate the stylistic alignment of the different models with the training set of each player. For each position contained in the test split, a single move is generated by the evaluated model. For Maia-2 and Maia-2 FT, the selected move corresponds to the action with the highest probability predicted by the policy network. For MCTS, the selected move corresponds to the most visited action at the end of the search procedure. The resulting board transitions induced by these generated moves are then projected into the latent evaluation space, discretized, and compared to the corresponding player transition distributions using the proposed JSD-based metric.

The results, shown in Table \ref{tab:jsd_stability}, indicate that the original Maia-2 model has a relatively high divergence compared to the training set of each player, with an average of $0.159$. This is expected because Maia-2 was not specifically trained to reproduce the behaviors of the champions included in this dataset. Similarly to the move accuracy evaluation, the fine-tuning process leads to an improvement in stylistic alignment, reducing the average divergence to $0.141$. The introduction of MCTS further improves stylistic alignment, with an average divergence of $0.101$, corresponding to an additional decrease of $0.04$ compared to Maia-2 FT.

These results contrast with the move accuracy evaluation, where the introduction of MCTS led to a significant decrease in deterministic performance. One possible explanation is that the search process biases move selection toward board transitions that are behaviorally closer to those observed in the target player's games, even when the exact played move differs from the ground truth. This observation highlights the limitation of move accuracy for evaluating style-oriented agents.

While these results provide promising evidence that the proposed Jensen-Shannon metric captures meaningful behavioral similarities between players and models, this approach should still be considered an initial exploration. In particular, part of the observed improvement may reflect distributional smoothing effects introduced by search and dimensionality reduction rather than stylistic fidelity alone. Further comparisons with additional behavioral metrics and qualitative analyses are therefore necessary to fully validate the robustness of this evaluation framework.

\begin{table}[!t]
	\renewcommand{\arraystretch}{1.3}
	\caption{Jensen-Shannon divergence compared to the training set for each player and each model. Values following the $\pm$ indicate the standard deviation.}
	\label{tab:jsd_stability}
	\centering
	\scriptsize
	\setlength{\tabcolsep}{3pt}
	\begin{tabular}{l c c c c}
		\hline
		\bfseries Player  & Test                      & Maia-2                    & Maia-2 FT                 & Maia-2 FT + MCTS          \\
		\hline\hline
		Alekhine          & $0.051\pm0.002$           & $0.175\pm0.002$           & $0.129\pm0.002$           & $0.100\pm0.002$           \\
		Anand             & $0.036\pm0.001$           & $0.148\pm0.001$           & $0.094\pm0.002$           & $0.093\pm0.002$           \\
		Andersson         & $0.048\pm0.002$           & $0.150\pm0.002$           & $0.149\pm0.002$           & $0.110\pm0.002$           \\
		Beliavsky         & $0.042\pm0.001$           & $0.170\pm0.001$           & $0.137\pm0.002$           & $0.107\pm0.002$           \\
		Capablanca        & $0.065\pm0.002$           & $0.172\pm0.002$           & $0.167\pm0.003$           & $0.099\pm0.003$           \\
		Fischer           & $0.069\pm0.002$           & $0.174\pm0.003$           & $0.176\pm0.003$           & $0.140\pm0.003$           \\
		Ivanchuk          & $0.037\pm0.001$           & $0.150\pm0.001$           & $0.129\pm0.001$           & $0.082\pm0.002$           \\
		Karpov            & $0.037\pm0.001$           & $0.161\pm0.001$           & $0.134\pm0.001$           & $0.104\pm0.002$           \\
		Kasparov          & $0.048\pm0.002$           & $0.163\pm0.002$           & $0.149\pm0.002$           & $0.092\pm0.002$           \\
		Korchnoi          & $0.042\pm0.001$           & $0.156\pm0.001$           & $0.132\pm0.001$           & $0.096\pm0.001$           \\
		Larsen            & $0.044\pm0.001$           & $0.129\pm0.002$           & $0.128\pm0.002$           & $0.077\pm0.002$           \\
		Petrosian         & $0.054\pm0.002$           & $0.164\pm0.002$           & $0.153\pm0.002$           & $0.105\pm0.002$           \\
		Portisch          & $0.041\pm0.001$           & $0.167\pm0.002$           & $0.139\pm0.002$           & $0.106\pm0.002$           \\
		Short             & $0.042\pm0.001$           & $0.159\pm0.002$           & $0.154\pm0.002$           & $0.109\pm0.002$           \\
		Tal               & $0.046\pm0.002$           & $0.148\pm0.002$           & $0.146\pm0.002$           & $0.096\pm0.002$           \\
		Timman            & $0.036\pm0.001$           & $0.152\pm0.001$           & $0.137\pm0.001$           & $0.093\pm0.002$           \\
		\hline
		\bfseries Average & \bfseries $0.046\pm0.002$ & \bfseries $0.159\pm0.002$ & \bfseries $0.141\pm0.002$ & \bfseries $0.101\pm0.002$ \\
		\hline
	\end{tabular}
\end{table}

\section{Conclusion}
This work proposes a pipeline combining individualized Maia-2 embeddings, MCTS-enhanced move generation, and a behavioral metric based on the Jensen-Shannon divergence to model player-specific chess behaviors. Results suggest that move accuracy alone is insufficient for evaluating stylistic alignment and that distribution-based metrics provide a more informative alternative.

Several promising directions remain for future work. First, the proposed JSD-based metric should be validated against additional behavioral indicators such as opening preferences, blunder rates, or other stylistic descriptors to further assess its robustness. Extending the evaluation to amateur players and other strategic games would also help verify the generalizability of the approach. Finally, future work could leverage the proposed framework to analyze the historical evolution of chess behaviors across different eras and player populations.

\section*{Acknowledgment}
\ifanonymous
\else
	This article is based upon work from COST Action CA22145 - GameTable, supported by COST (European Cooperation in Science and Technology). The Ludus ex Machina project is supported by Schmidt Sciences (grant number 25-70261). Computational resources have been provided by the Consortium des Equipements de Calcul Intensif (CECI), funded by the Fonds de la Recherche Scientifique de Belgique (F.R.S.-FNRS) under Grant No. 2.5020.11 and by the Walloon Region.
\fi

\bibliographystyle{IEEEtran}
\bibliography{references}

\end{document}